\documentclass[letterpaper]{article}
\usepackage[preprint]{aaai2027}

\usepackage[hyphens]{url}
\usepackage{graphicx}
\usepackage{natbib}
\usepackage{caption}
\usepackage{algorithm}
\usepackage{algorithmic}
\usepackage{pifont}

\usepackage{tabularx}
\usepackage{url}
\usepackage{booktabs}
\usepackage{colortbl}
\usepackage{siunitx}
\usepackage{multirow}
\usepackage{makecell}
\definecolor{proposedbg}{HTML}{E0EAF5}
\usepackage{amsmath}
\usepackage{amsfonts}
\usepackage{amssymb}
\usepackage{nicefrac}
\usepackage{microtype}

\usepackage{newfloat}
\usepackage{listings}
\DeclareCaptionStyle{ruled}{labelfont=normalfont,labelsep=colon,strut=off}
\floatstyle{ruled}
\newfloat{listing}{tb}{lst}{}
\floatname{listing}{Listing}

\usepackage{booktabs}
\usepackage{amsmath}
\usepackage{amssymb}

\title{Deep Thought Alignment: Trajectory-Level Latent Distillation for Video Reasoning}
\author{
    Ao Shen\textsuperscript{\rm 1,2$\ast$},
    Yongheng Zhang\textsuperscript{\rm 1}\thanks{~Equal Contribution. $^\dagger$ Corresponding Author.},
    Yinghui Li\textsuperscript{\rm 1}$^\dagger$,
    Manning Wang\textsuperscript{\rm 2}$^\dagger$,
    Di Yin\textsuperscript{\rm 1},
    Xing Sun\textsuperscript{\rm 1}
}
\affiliations{
    \textsuperscript{\rm 1}Tencent Youtu Lab \ \ \  \textsuperscript{\rm 2}Fudan University\\
    lebronyhli@tencent.com
    \vspace{-10mm}
}

\begin{document}

\maketitle

\vspace*{-4em}

\begin{abstract}
Large Multimodal Models (LMMs) for video reasoning have long been hindered by the high computational cost of processing vast amounts of visual information. This dilemma motivates the transfer of the reasoning capabilities of large models to smaller, more efficient ones. On-Policy Distillation (OPD) offers a promising solution by matching output-token distributions along student-generated trajectories. However, video reasoning often depends on evidence accumulated across multiple frames. In this context, output-level supervision only captures information expressed through token predictions and does not directly constrain the latent representations formed during reasoning. To address this limitation, we propose \textsc{Latent-OPD}, which augments OPD with trajectory-level latent distillation. Specifically, our method focuses on the position at the end of each trajectory, where hidden states effectively summarize the accumulated visual evidence and reasoning context. Furthermore, we introduce a progressive teacher-lookahead strategy, which aligns middle-to-late student layers with increasingly deeper teacher layers. Experiments on six video reasoning benchmarks show that \textsc{Latent-OPD} consistently outperforms output-only OPD. Notably, the improvements are particularly pronounced in scenarios with limited frames, long videos, or tasks requiring complex evidence aggregation. These results establish \textsc{Latent-OPD} as a highly effective approach to frame-efficient video reasoning.

\end{abstract}

\section{Introduction}

\begin{center}
{\large\itshape We know more than we can tell.}\\[0.65em]
\hfill{\small\itshape --- Michael Polanyi}
\end{center}

Video reasoning is a critical yet challenging capability in multimodal perception, underpinning temporal event understanding, spatial reasoning, and knowledge acquisition from dynamic visual streams~\cite{zhang2025vitcot,zhang2026thinkingvideovideogenerators,kuang2025natural,an2026toward,li2026cognitive}. Unlike single-frame image understanding, it requires models to process vast amounts of spatiotemporal information and distill sparse but decisive evidence for multi-step decisions. While recent Large Multimodal Models (LMMs) have substantially advanced this capability through stronger visual-language alignment and generative reasoning~\cite{team2026kimi,qwen3,lu2025youtu,zhang-etal-2024-wrong,zhang2026chatbotdigitalcolleagueparadigm}, their immense computational cost during inference severely limits their practical deployment. This dilemma creates a pressing need to transfer the reasoning capabilities of these massive models into smaller, more efficient counterparts.

\begin{figure}[t]
\centering
\includegraphics[width=1\columnwidth]{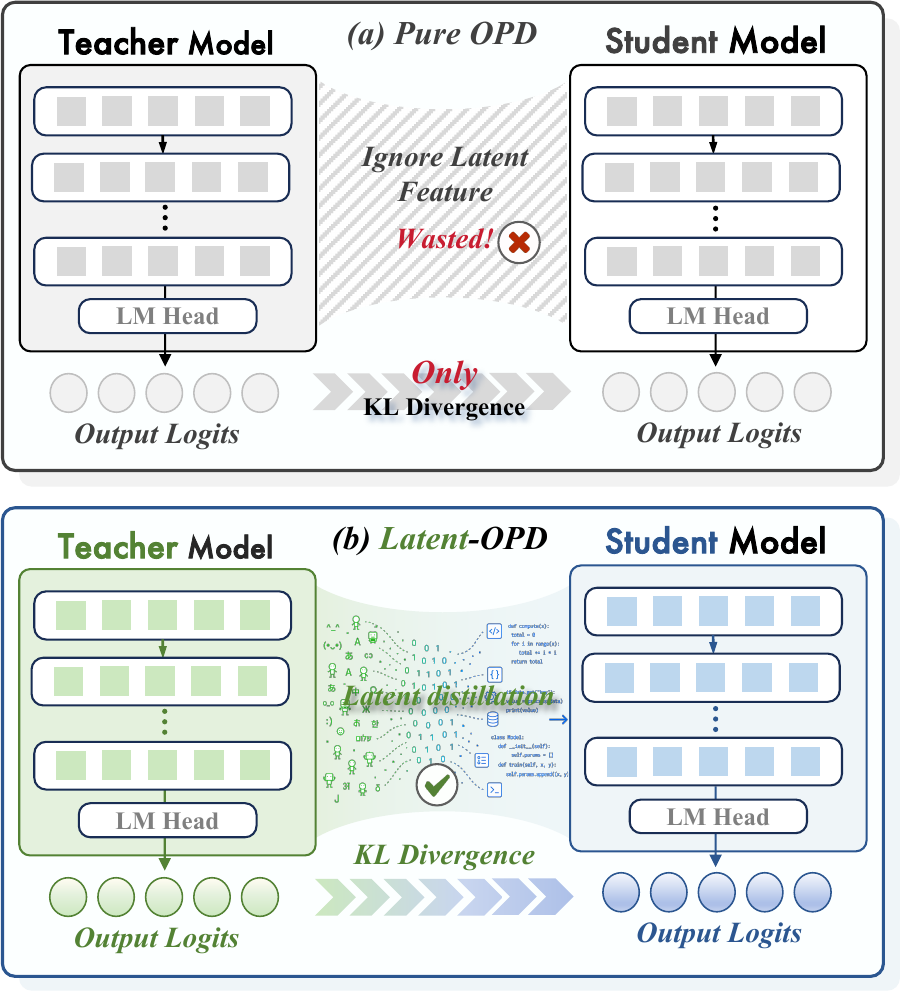}
\caption{Conceptual comparison between vanilla OPD and Latent-OPD. (a) Vanilla OPD transfers supervision only through output-level KL divergence, leaving the teacher's latent features unutilized. (b) Latent-OPD complements output-level KL divergence with latent distillation, aligning compact hidden states to transfer internal reasoning representations.}
\label{fig:intro}
\vspace{-1em}
\end{figure}

To achieve this capability transfer, the training paradigm has shifted from static, outcome-level imitation toward trajectory-level optimization. On-Policy Distillation (OPD) has emerged as a highly promising solution, aligning the student's output-token distributions with the teacher's along the student's self-generated trajectories~\cite{opd,vicur}. In visual domains, recent studies have begun to adapt OPD to different forms of visual supervision. \citet{visionopd} introduce Vision-OPD, extending this idea to fine-grained image understanding through regional-to-global on-policy self-distillation. \citet{liu2026visualadvantageonpolicydistillationvisionlanguage} propose VA-OPD, which identifies vision-critical tokens by comparing teacher scoring with and without fine-grained visual detail and reweights rollout- and token-level distillation accordingly.  \citet{videoopd} propose Video-OPD, applying on-policy dense supervision to semantic modeling in temporal video.

However, existing OPD methods for video reasoning rest on an overly optimistic assumption: they presuppose that complex spatiotemporal evidence can be fully captured solely by the output vocabulary distribution. As illustrated in Figure~\ref{fig:intro}(a), vanilla OPD operates exclusively at the output layer, supervising the student model's generated trajectory by minimizing KL divergence across output distributions. Consequently, it neglects the rich latent knowledge embedded in the teacher model's hidden representations preceding the language modeling head. This limitation echoes the tacit-knowledge view highlighted at the beginning of this section: token predictions may reveal only a fraction of the internal representations supporting them. In video reasoning, these latent states encode crucial spatiotemporal evidence and reasoning contexts along the generated trajectory. This motivates ``deep thought alignment'', which extends OPD from output distributions to trajectory-level hidden representations. While distilling these hidden states effectively transfers video reasoning capabilities, it introduces \textit{\textbf{two key challenges}}. \ding{172}~First, dense visual tokens contain substantial background noise and temporal redundancy, rendering token-level alignment costly and noisy. \ding{173}~Second, architectural differences between teacher and student models hamper direct layer-wise alignment. These challenges raise a central question: \textit{``How can trajectory-level hidden representations be efficiently transferred across heterogeneous models without dense visual-token alignment?''}

To address this question, we introduce Latent-space On-Policy Distillation (\textsc{\textbf{Latent-OPD}}), a trajectory-level latent distillation framework for deep thought alignment in video reasoning. Specifically, as illustrated in Figure~\ref{fig:intro}(b), \textsc{Latent-OPD} complements the standard output-level KL objective with a latent distillation branch. Instead of densely aligning all visual or response tokens, we extract tail hidden states from trajectories generated by the teacher as compact global anchors that integrate both video evidence and reasoning context. Furthermore, we employ a progressive teacher-lookahead mapping to align the student's middle-to-late layers with deeper teacher layers, while preserving the student's top layers for token-policy optimization and generation.

Across six comprehensive video reasoning benchmarks, \textsc{Latent-OPD} consistently outperforms vanilla OPD. These improvements are particularly pronounced in low- and mid-frame regimes, complex long-video scenarios, and demanding reasoning tasks requiring cross-frame evidence aggregation, such as action reasoning, spatial reasoning, and temporal counting. Notably, even with fewer input frames, \textsc{Latent-OPD} matches or surpasses the higher-frame vanilla OPD baseline, indicating a more efficient utilization of visual evidence rather than a mere reliance on high frame density.

Our main contributions are threefold:
\begin{itemize}
    \item[\ding{172}] We highlight the output-level bottleneck of vanilla OPD for video reasoning: final-logit supervision improves token preferences but underexploits latent spatiotemporal representations before the language modeling head.
    \item[\ding{173}] We introduce \textsc{Latent-OPD}, which performs deep thought alignment using sparse trajectory-tail anchors and progressive teacher lookahead, enabling efficient latent transfer across heterogeneous architectures.
    \item[\ding{174}] We conduct extensive evaluations across six video reasoning benchmarks, showing consistent improvements over vanilla OPD and clear advantages in frame efficiency and long-video understanding scenarios.
\end{itemize}

\begin{figure*}[t]
\centering
\includegraphics[width=0.95\textwidth]{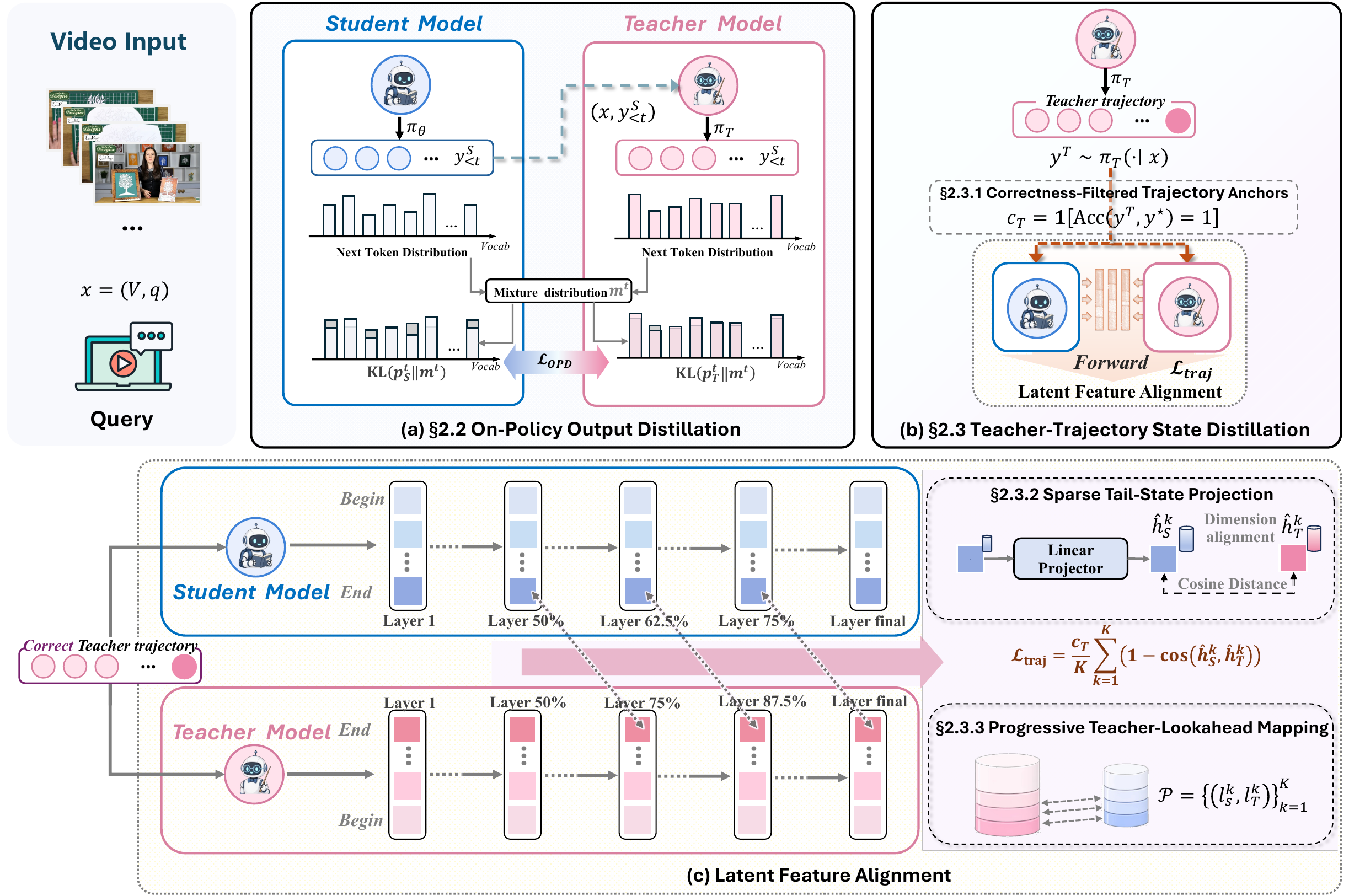}

\caption{Overview of \textsc{Latent-OPD}.
(a) On-Policy Output Distillation: supervises student-sampled trajectories using teacher token distributions.
(b) Teacher-Trajectory State Distillation: selects correct teacher rollouts as latent anchors.
(c) Latent Feature Alignment: aligns projected student states to teacher states at selected layers via cosine distance.
}
\label{fig:framework}

\end{figure*}

\section{Method}
We present Latent-space On-Policy Distillation (\textsc{Latent-OPD}), a trajectory-level distillation framework that complements output-token preference transfer with latent-state supervision from a frozen teacher, enabling deep thought alignment for complex video reasoning in LMMs.

\subsection{Overview}
\label{sec:overview}

Given a video-question input $x=(V,q)$, \textsc{Latent-OPD} trains a student policy $\pi_\theta$ with a frozen teacher $\pi_T$. In the output stream, the student samples $y^S \sim \pi_\theta(\cdot \mid x)$, and the teacher supplies token distributions on the same student-visited prefixes $(x,y^S_{<t})$ (Figure~\ref{fig:framework}(a)). In the latent stream, the teacher generates $y^T \sim \pi_T(\cdot \mid x)$, the generated trajectories are filtered for reliability to serve as latent anchors, and $z=[x;y^T]$ is fed to both models under teacher forcing. We then align only the final valid-token hidden states across selected layer pairs, avoiding dense hidden-state matching over all video and response tokens (Figure~\ref{fig:framework}(b--c)). Crucially, the teacher, filter, and projection heads are used only during training; the inference phase remains strictly student-only.

\subsection{On-Policy Output Distillation}
\label{sec:opd}
As shown in Figure~\ref{fig:framework}(a), given an input $x$, the student samples a response trajectory $y^S \sim \pi_\theta(\cdot \mid x)$ of length $L_S$. Then, at each decoding step $t$, both the student and the frozen teacher are evaluated on the same prefix $(x,y^S_{<t})$, producing the next-token distributions
$p_S^t=\pi_\theta(\cdot \mid x,y^S_{<t})$ and
$p_T^t=\pi_T(\cdot \mid x,y^S_{<t})$, respectively. With these paired distributions, we optimize a weighted JSD-style objective to obtain bounded and stable token-level supervision:

\begin{equation}
\mathcal{L}_{\mathrm{OPD}}
=
\frac{1}{L_S}
\sum_{t=1}^{L_S}
\left[
(1-\alpha)\mathrm{KL}(p_S^t \| m^t)
+
\alpha\mathrm{KL}(p_T^t \| m^t)
\right],
\label{eq:opd}
\end{equation}

where $m^t = (1-\alpha)p_S^t + \alpha p_T^t$ is the interpolated mixture distribution, and $\alpha\in(0,1)$ controls the teacher-side weight. The teacher and student share the same tokenizer.

For efficiency, we approximate this objective on the student's top-$k$ token support plus a shared tail bucket. Specifically, the residual student and teacher probability masses outside the top-$k$ support are both accumulated into this bucket, preserving the non-top-$k$ density without artificially renormalizing the truncated vocabulary.

\subsection{Teacher-Trajectory State Distillation}
\label{sec:trajectory_state}

Since output distillation only supervises post-softmax behavior, the teacher's internal multimodal evidence aggregation remains largely untransferred. To bridge this gap, we introduce a teacher-trajectory state distillation branch via three components: correctness-filtered trajectory anchors, sparse tail-state projection, and progressive teacher-lookahead mapping.

\subsubsection{Correctness-Filtered Trajectory Anchors}
\label{sec:correct_anchor}
\begin{table*}[t]
\centering
\footnotesize
\setlength{\tabcolsep}{4.5pt}
\renewcommand{\arraystretch}{1.12}

\resizebox{\textwidth}{!}{
\begin{tabular}{l|c|cccccc|c}
\toprule[1.0pt]

\textbf{Model} & \textbf{Frames} & \textbf{VSI-Bench} & \textbf{Video-MMMU} & \textbf{MMVU} & \textbf{MVBench} & \textbf{TempCompass} & \textbf{Video-MME} & \textbf{Avg.} \\
\midrule

LongVA-7B$^\dagger$ & -- & 29.2 & 23.9 & -- & -- & 56.9 & 52.6 & -- \\
VILA-1.5-8B$^\dagger$ & -- & 28.9 & 20.8 & -- & -- & 58.8 & -- & -- \\
VILA-1.5-40B$^\dagger$ & -- & 31.2 & 34.0 & -- & -- & -- & 60.1 & -- \\
Video-UTR-7B$^\dagger$ & -- & -- & -- & -- & 58.8 & 59.7 & 52.6 & -- \\
LLaVA-OneVision-7B$^\dagger$ & -- & 32.4 & 33.8 & 49.2 & 56.7 & -- & 58.2 & -- \\
Kangaroo-8B$^\dagger$ & -- & -- & -- & -- & 61.1 & 62.5 & 56.0 & -- \\
Video-R1-7B$^\dagger$ & 16 & 34.6 & 49.8 & 64.2 & 62.7 & 72.6 & 57.4 & 56.9 \\
Video-R1-7B$^\dagger$ & 32 & 35.8 & 52.3 & 63.8 & 63.9 & 73.2 & 59.3 & 58.1 \\
Video-R1-7B$^\dagger$ & 64 & 37.1 & 52.4 & 63.8 & 64.8 & 73.2 & 61.4 & 58.8 \\
\midrule
Qwen3.5-9B-CoT & 16 & 29.4 & 41.2 & 61.6 & 51.8 & 66.6 & 51.0 & 50.3 \\
Qwen3.5-9B-CoT & 32 & 33.4 & 42.8 & 59.8 & 52.2 & 66.5 & 50.3 & 50.8 \\
Qwen3.5-9B-CoT & 64 & 27.9 & 44.0 & 61.6 & 52.0 & 66.5 & 48.5 & 50.1 \\
\midrule
Qwen3.5-9B-SFT+GRPO & 16 & 47.2 & 59.8 & 65.9 & 60.7 & 72.2 & 60.3 & 61.0 \\
Qwen3.5-9B-SFT+GRPO & 32 & 50.0 & 60.3 & 66.6 & 60.7 & 72.2 & 61.6 & 61.9 \\
Qwen3.5-9B-SFT+GRPO & 64 & 52.6 & 59.7 & 67.2 & 60.8 & \textbf{72.2} & 64.6 & 62.9 \\
\midrule
Vanilla OPD-9B & 16 & 47.0 & 61.0 & 70.6 & 64.7 & 72.2 & 59.2 & 62.5 \\
Vanilla OPD-9B & 32 & 48.8 & 60.4 & 71.7 & 65.0 & 71.6 & 60.7 & 63.0 \\
Vanilla OPD-9B & 64 & 51.9 & 64.4 & 73.6 & 64.5 & 70.5 & 64.7 & 64.9 \\
\midrule
\rowcolor{gray!15} \textsc{Latent-OPD}-9B & 16 & \textbf{48.2} & \textbf{65.4} & \textbf{73.3} & \textbf{64.9} & \textbf{73.2} & \textbf{61.6} & \textbf{64.4}$_{\uparrow 1.9}$ \\
\rowcolor{gray!15} \textsc{Latent-OPD}-9B & 32 & \textbf{51.7} & \textbf{67.4} & \textbf{72.6} & \textbf{65.4} & \textbf{72.4} & \textbf{64.3} & \textbf{65.6}$_{\uparrow 2.6}$ \\
\rowcolor{gray!15} \textsc{Latent-OPD}-9B & 64 & \textbf{54.9} & \textbf{67.2} & \textbf{74.4} & \textbf{65.1} & 70.6 & \textbf{66.5} & \textbf{66.5}$_{\uparrow 1.6}$ \\
\bottomrule[1.0pt]
\end{tabular}
}

\caption{Accuracy (\%) on six benchmarks under 16-, 32-, and 64-frame budgets. $^\dagger$ denotes results reported in prior work; bold marks the best Qwen3.5-9B result for each benchmark and frame budget. Subscripts denote the improvement over vanilla OPD.}
\label{tab:main-results}

\end{table*}
We first decide which teacher trajectories should provide latent supervision. As shown in Figure~\ref{fig:framework}(b), for each input $x$, the teacher independently generates a complete reasoning trajectory $y^T \sim \pi_T(\cdot \mid x)$ from the original video-question input. Since plausible rationales can still lead to wrong answers, we use final-answer correctness as a lightweight reliability proxy: for ground-truth answer $y^\star$, $c_T=\mathbf{1}[\operatorname{Acc}(y^T,y^\star)=1]$, where $\operatorname{Acc}$ denotes the task evaluator applied to the parsed final answer; unlabeled examples are excluded from the training set. After this filtering step, the retained sequence $z=[x;y^T]$ is used as the shared input for hidden-state alignment.

\subsubsection{Sparse Tail-State Projection}
\label{sec:tail_alignment}

Given the retained anchors, we next localize where hidden-state alignment should occur. As shown in Figure~\ref{fig:framework}(c), for a retained sequence $z=[x;y^T]$, we perform parallel forward passes through the frozen teacher and the trainable student. Let $e$ denote the index of the final valid token of the teacher trajectory in $z$, e.g., the last non-padding response token. Because a causal decoder's tail state can attend to the full video-question context and preceding reasoning tokens, it serves as a compact trajectory summary. We therefore align hidden states only at this position, avoiding dense token-level matching over the full multimodal sequence.

Then, for each selected layer pair $(l_S^k,l_T^k)$, we extract the student and teacher hidden states at position $e$. Since the two models may have different hidden dimensions, the student state is mapped into the teacher's latent space using a trainable linear projection head $P_k$. We then apply $\ell_2$ normalization:
\begin{equation}
\hat{h}_S^k
=
\operatorname{norm}\left(P_k \left(h_\theta^{l_S^k}(z)_e\right)\right),
\qquad
\hat{h}_T^k
=
\operatorname{norm}\left(h_T^{l_T^k}(z)_e\right),
\end{equation}
where $\operatorname{norm}(u)=u/(\|u\|_2+\epsilon)$ for a small $\epsilon>0$. The trajectory-level latent objective minimizes the average cosine distance over $K$ selected layer pairs:
\begin{equation}
\mathcal{L}_{\mathrm{traj}}
=
\frac{c_T}{K}
\sum_{k=1}^{K}
\left(
1-\cos\left(\hat{h}_S^k,\hat{h}_T^k\right)
\right).
\label{eq:global_alignment}
\end{equation}
Here, $c_T$ acts as a binary mask for unreliable trajectories. For each minibatch, we average this loss over retained examples only and set $\mathcal{L}_{\mathrm{traj}}=0$ if none are retained.

\subsubsection{Progressive Teacher-Lookahead Mapping}
\label{sec:lookahead_mapping}

After choosing the tail state as the alignment target, we still need to decide which student and teacher layers should be paired. To construct the layer pairs in Equation~\ref{eq:global_alignment}, we employ an asymmetric teacher-lookahead mapping rather than matching identical relative depths. Let $N_S$ and $N_T$ be the numbers of student and teacher layers under 1-based block indexing, and denote the selected pairs by $\mathcal{P}=\{(l_S^k,l_T^k)\}_{k=1}^{K}$, where $l_S^k=1+\lfloor r_S^k (N_S-1)\rfloor$ and $l_T^k=1+\lfloor r_T^k (N_T-1)\rfloor$. We choose the relative depths so that $r_S^k<r_T^k$ and the resulting discrete indices are unique and satisfy $l_S^1<\cdots<l_S^K$ and $l_T^1<\cdots<l_T^K$, forming a progressive mapping in which each selected student layer looks ahead to a deeper teacher layer. Consequently, the student's middle-to-late layers, excluding the final ones, are exposed to more abstract teacher representations, while the output-facing student layers remain unconstrained by the latent loss. The exact layer configurations are detailed in the experimental setup.

\begin{table}[h]
\centering
\footnotesize
\setlength{\tabcolsep}{3.0pt}
\renewcommand{\arraystretch}{1.08}

\begin{tabular}{>{\raggedright\arraybackslash}m{0.68\columnwidth}|>{\centering\arraybackslash\scriptsize}m{0.065\columnwidth}>{\centering\arraybackslash\scriptsize}m{0.065\columnwidth}>{\centering\arraybackslash\scriptsize}m{0.065\columnwidth}}
\toprule[1.0pt]
\textbf{Variant} & \textbf{16f} & \textbf{32f} & \textbf{64f} \\
\midrule
 \mbox{Vanilla OPD} & 59.20 & 60.70 & 64.70 \\
\midrule
\multicolumn{4}{l}{\textit{Controlled layer mapping}} \\
\midrule
Same-depth, fixed student layers & 59.81 & 63.85 & 65.11 \\
 Fixed-offset $(+12.5\%)$ & 59.89 & 62.81 & 65.30 \\
Reverse-lookahead & 59.89 & 63.33 & 65.93 \\
Single tail pair ($s_{75\%}\!\to\!t_{100\%}$) & 59.60 & 64.00 & 65.70 \\
 $+$ final pair & 58.90 & 63.30 & 65.00 \\
$+$ early pair & 59.90 & 63.20 & 65.40 \\
\midrule
\multicolumn{4}{l}{\textit{Representation / token-level controls}} \\
\midrule
 OPRD-style distill & 59.70 & 63.85 & 65.78 \\
Dense token-level hidden KD & 59.20 & 63.10 & 66.10 \\
 Low-rank projector $(r=16)$ & 58.60 & 63.30 & 65.00 \\
Teacher-trajectory SFT  & 59.96 & 63.78 & 65.70 \\
 Reverse-KL output distillation & 60.60 & 62.90 & 65.70 \\
\midrule
\multicolumn{4}{l}{\textit{Trajectory source and anchor}} \\
\midrule
w/o correctness-filtered trajectory anchors & 59.10 & 63.50 & 65.10 \\
Independent teacher/student trajectories & 60.30 & 63.40 & 65.20 \\
Shared student-generated trajectory & 60.40 & 62.90 & 66.00 \\
Prompt-end latent anchor & 59.60 & 62.90 & 64.90 \\
 Answer-end latent anchor & 60.50 & 62.90 & 65.80 \\
Think-end latent anchor & 60.22 & 63.67 & 65.22 \\
\midrule
 \textsc{Latent-OPD} (ours) & \textbf{61.60} & \textbf{64.30} & \textbf{66.50} \\
\bottomrule[1.0pt]
\end{tabular}

\caption{Ablation study on the Video-MME benchmark evaluating layer mapping, representation- and token-level supervision, trajectory source, and anchor position. }
\label{tab:ablation}

\end{table}

\subsection{Training Objective and Inference}

\label{sec:training_inference}

The on-policy output distillation and teacher-trajectory state distillation streams are jointly optimized through a unified objective. Specifically, the generative component anchors the student's output distribution, while the latent component aligns its intermediate states. The total loss is defined as:
\begin{equation}
\begin{aligned}
\mathcal{L}_{\mathrm{total}}
&=\mathcal{L}_{\mathrm{gen}}
+\operatorname{clip}_{\rho\operatorname{sg}(\lvert\mathcal{L}_{\mathrm{gen}}\rvert)}
\left(\lambda_g\omega(\tau)\mathcal{L}_{\mathrm{traj}}\right),\\
\mathcal{L}_{\mathrm{gen}}
&=\mathcal{L}_{\mathrm{OPD}}
+\beta\mathcal{L}_{\mathrm{refKL}}
+\lambda_{\mathrm{fmt}}\mathcal{L}_{\mathrm{format}},
\end{aligned}
\label{eq:total_loss}
\end{equation}
where the generative loss $\mathcal{L}_{\mathrm{gen}}$ aggregates the core OPD supervision $\mathcal{L}_{\mathrm{OPD}}$, a reference-policy KL penalty $\mathcal{L}_{\mathrm{refKL}}$, and an answer-format loss $\mathcal{L}_{\mathrm{format}}$ (weighted by $\beta$ and $\lambda_{\mathrm{fmt}}$, respectively). The latent trajectory loss $\mathcal{L}_{\mathrm{traj}}$ is regulated by two mechanisms: a linear warmup factor $\omega(\tau) = \min(1, \tau/\tau_w)$ over the first $\tau_w$ steps, and a dynamic clipping function $\operatorname{clip}_{b}(u) = u \cdot \min(1, b/(\lvert u\rvert+\epsilon))$, for a small $\epsilon>0$. Here, $\operatorname{sg}(\cdot)$ detaches the clipping budget; thus this clipping acts as a scalar contribution cap rather than a gradient-norm constraint.

During inference, the teacher model and projection heads are no longer required, leaving the deployed student with no additional computational overhead.

\section{Experiments}

\subsection{Experimental Setting}
\label{sec:experimental-setting}

\textbf{Implementation Details:}
Following Video-R1~\cite{videor1}, our student backbone (Qwen3.5-9B-Base~\cite{qwen35}) undergoes SFT on the Video-R1 CoT dataset, followed by \textsc{Latent-OPD} training on its RL dataset. As the teacher, we use a frozen Qwen3.5-27B-Base~\cite{qwen35} video CoT model that was previously fine-tuned. In the latent branch, we align the final valid-token states at layer mappings $(s_{50\%}\!\to\!t_{75\%})$, $(s_{62.5\%}\!\to\!t_{87.5\%})$, and $(s_{75\%}\!\to\!t_{100\%})$. We set the latent loss weight to 0.01, incorporating a 5\% warmup schedule and a cap at 15\% of the generative loss. Consistent with Video-R1, video preprocessing involves sampling training clips at 1 FPS and tokenizing visual inputs with a $28 \times 28$ patch size. During evaluation, we test 16, 32, and 64 frames.

\textbf{Benchmarks and Baselines:}
Following the Video-R1 evaluation protocol, we report accuracy on six public video reasoning benchmarks: VSI-Bench~\cite{vsibench}, Video-MMMU~\cite{videommmu}, MMVU~\cite{mmvu}, MVBench~\cite{mvbench}, TempCompass~\cite{tempcompass}, and Video-MME~\cite{videomme}. The compared methods include representative prior video LMMs, namely  LongVA-7B~\cite{longva}, VILA-1.5-8B/40B~\cite{vila}, Video-UTR-7B~\cite{videoutr}, LLaVA-OneVision-7B~\cite{llavaonevision}, Kangaroo-8B~\cite{kangaroo}, and Video-R1-7B~\cite{videor1}. We additionally evaluate CoT, SFT+GRPO, Vanilla OPD, and \textsc{Latent-OPD} on Qwen3.5-9B-Base, where vanilla OPD removes only the latent trajectory branch of \textsc{Latent-OPD}. Further implementation and baseline details are provided in Appendix A.

\subsection{Main Results}
\label{sec:main-results}
\begin{figure}[t]
\centering
\includegraphics[width=0.98\columnwidth]{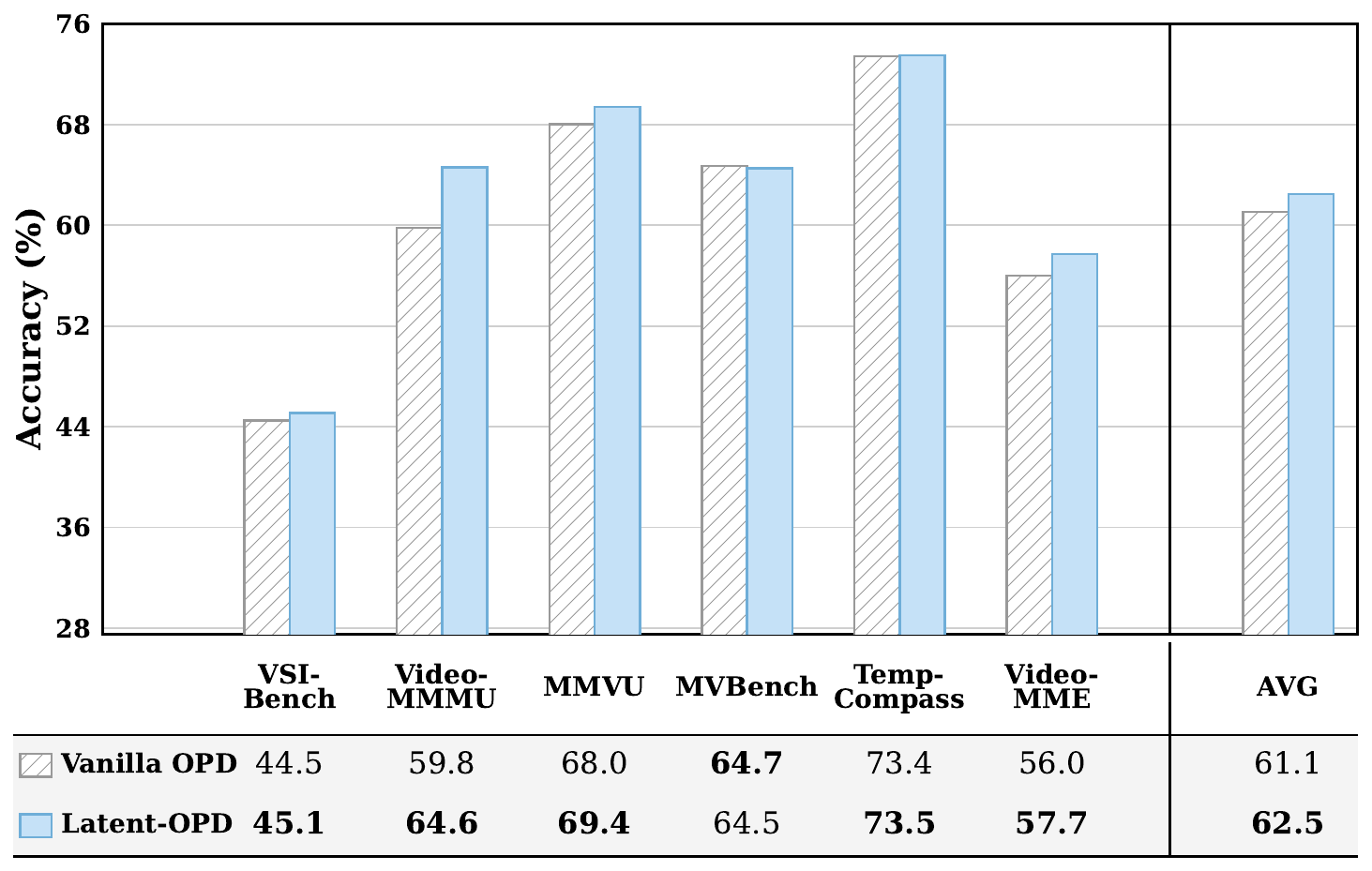}
\caption{Accuracy (\%) of vanilla OPD and \textsc{Latent-OPD} with a Qwen3.5-4B student under the 16-frame setting. The final group reports the six-benchmark average.}
\label{fig:4b-f16}

\end{figure}

\begin{figure}[t]
\centering
\includegraphics[width=0.9\columnwidth]{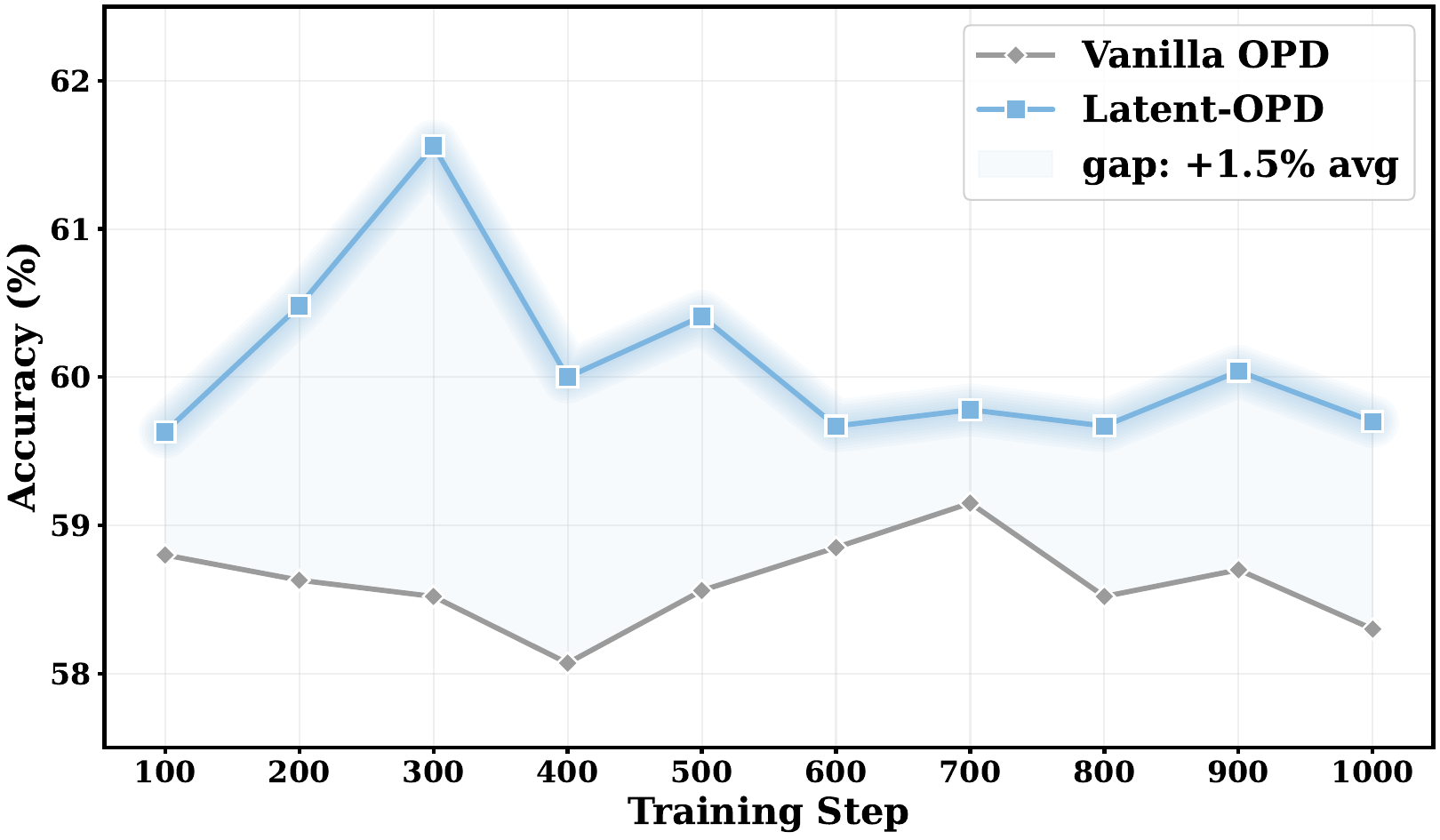}
\caption{Video-MME accuracy (\%) during an extended 1,000-step diagnostic under the 16-frame setting.}
\label{fig:training-dynamics}

\end{figure}

Table~\ref{tab:main-results} presents the main experimental results. From this table, we can make three key observations:

\textbf{Obs. 1. The full post-training pipeline delivers consistent, efficient, and stable gains.} \textsc{Latent-OPD} outperforms Qwen3.5-9B-CoT on all 18 benchmark/frame-budget pairs, raising the six-benchmark average by $14.2$, $14.8$, and $16.4$ points at 16, 32, and 64 frames, respectively. It also surpasses Qwen3.5-9B-SFT+GRPO by $3.4$, $3.7$, and $3.6$ average points and outperforms vanilla OPD by $2.0$, $2.6$, and $1.5$ average points under the same frame budgets. Beyond final accuracy, \textsc{Latent-OPD} converges faster in terms of training steps: as shown in Figure~\ref{fig:training-dynamics}, the diagnostic curves separate from vanilla OPD early and remain stable. These results confirm that the proposed latent distillation supplies informative optimization signals while scaling robustly with extended visual contexts.

\begin{figure}[t]
\centering
\includegraphics[width=0.98\columnwidth]{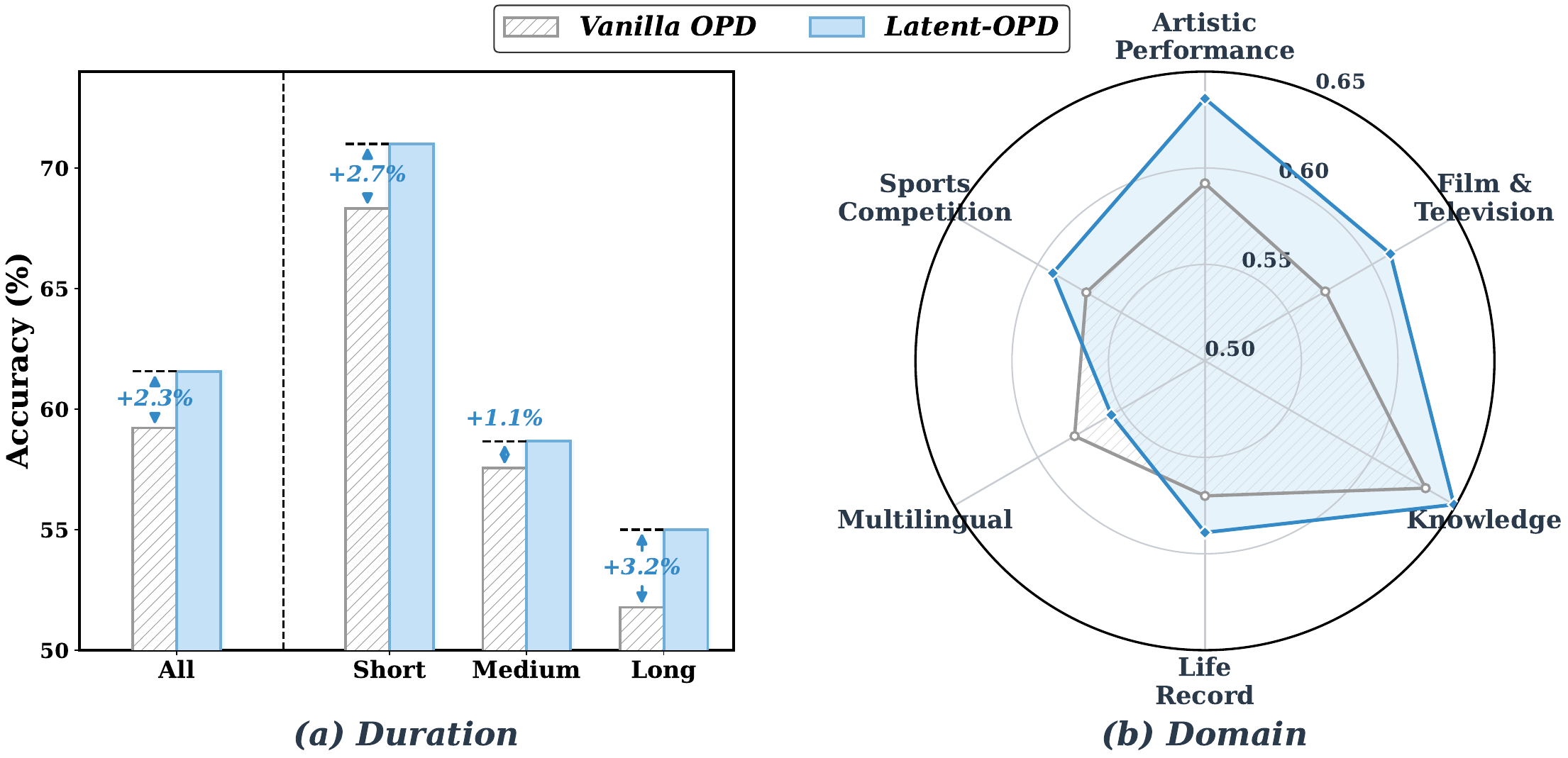}
\caption{Video-MME accuracy (\%) under the 16-frame setting, broken down by (a) video duration and (b) domain.}
\label{fig:finegrained-f16}
\vspace{-1em}
\end{figure}

\textbf{Obs. 2. \textsc{Latent-OPD} excels at long-horizon temporal integration.} Compared to vanilla OPD, the most significant gains emerge on tasks requiring cross-frame reasoning, notably Video-MMMU (up to $+7.0$) and Video-MME (up to $+3.6$). As detailed in the 16-frame Video-MME breakdown (Figure~\ref{fig:finegrained-f16}), improvements peak on long videos ($+3.22$) and temporally complex domains like Artistic Performance ($+4.44$) and Film \& Television ($+3.89$). This outcome validates the essential role of trajectory-tail alignment in successfully consolidating distributed visual evidence.

\textbf{Obs. 3. Gains extend seamlessly to frame efficiency.} \textsc{Latent-OPD} extracts richer signals per frame, allowing its 16-frame average ($64.4\%$) to comfortably surpass vanilla OPD's 32-frame performance ($63.0\%$). Similarly, the 32-frame \textsc{Latent-OPD} matches or beats the 64-frame baseline on challenging sets like Video-MMMU, MVBench, and TempCompass, showing that latent trajectory alignment improves not only accuracy but also frame utilization efficiency.

\subsection{Performance on the 4B Student}
Beyond the 9B setting, we further evaluate \textsc{Latent-OPD} on a smaller Qwen3.5-4B-Base \cite{qwen35} student under the same six-benchmark evaluation protocol. As illustrated in Figure~\ref{fig:4b-f16}, \textsc{Latent-OPD} boosts the 16-frame average from $61.1\%$ to $62.5\%$. The gains are especially visible on reasoning-intensive benchmarks such as Video-MMMU, MMVU, and Video-MME. This observation suggests that the trajectory-level latent signals are highly informative, successfully transferring complex reasoning capabilities even when the student model's parameter capacity is significantly reduced.

\subsection{Ablation Study}
\label{sec:ablation-study}

Table~\ref{tab:ablation} ablates three key components on Video-MME:

\textbf{Controlled layer mapping:} To isolate teacher lookahead, all three-pair controls fix the student depths at $50\%$, $62.5\%$, and $75\%$ and vary only teacher depth. The default uses $(s_{50\%}\!\to\!t_{75\%})$, $(s_{62.5\%}\!\to\!t_{87.5\%})$, and $(s_{75\%}\!\to\!t_{100\%})$; same-depth uses $(s_{50\%}\!\to\!t_{50\%})$, $(s_{62.5\%}\!\to\!t_{62.5\%})$, and $(s_{75\%}\!\to\!t_{75\%})$; fixed-offset uses $(s_{50\%}\!\to\!t_{62.5\%})$, $(s_{62.5\%}\!\to\!t_{75\%})$, and $(s_{75\%}\!\to\!t_{87.5\%})$; reverse-lookahead uses $(s_{50\%}\!\to\!t_{25\%})$, $(s_{62.5\%}\!\to\!t_{50\%})$, and $(s_{75\%}\!\to\!t_{62.5\%})$. Pair-count controls keep only $(s_{75\%}\!\to\!t_{100\%})$, add $(s_{100\%}\!\to\!t_{100\%})$, or add $(s_{25\%}\!\to\!t_{50\%})$ to the default. Across the 16, 32, and 64 frame settings, the default mapping uniformly surpasses all variants, showing that mid-to-late student depths benefit most from deeper teacher states while output-adjacent depths remain dedicated to token-policy optimization.

\textbf{Representation and token-level controls:} We evaluate several variants to systematically validate our design choices. First, to verify our representation alignment strategy, we compare against dense token-level hidden knowledge distillation (KD) and an OPRD-style baseline~\cite{oprd} that densely matches response-token representations on student rollouts using same-depth pairs and normalized MSE. Both variants consistently underperform across all frame budgets, confirming that our selective state matching is superior to dense feature matching. Second, replacing our default linear mapping with a low-rank projector yields sub-optimal results, clearly validating the rationality of our full projector design. Third, replacing our output-level objective with Reverse-KL distillation reduces performance, demonstrating the clear advantage of our JSD-style divergence. Finally, standard token-level SFT on the same correct teacher trajectories underperforms, proving our gains stem from utilizing latent states rather than mere exposure to high-quality rationales.

\textbf{Trajectory source and anchor:} We justify our choices for the alignment source and anchor position. First, using unfiltered trajectory anchors degrades performance, confirming the need for correctness filtering. For the trajectory source, aligning tail states from independently generated student and teacher rollouts proves sub-optimal due to mismatched reasoning paths. Conversely, passing the same student-generated trajectory through both models aligns token positions but fails to leverage the teacher's superior reasoning. \textsc{Latent-OPD} resolves this by feeding both models the identical correctness-filtered teacher trajectory, ensuring a position-matched target that reflects the teacher's preferred logic. Finally, shifting the supervision anchor from our default final valid state to other boundaries (prompt-end, answer-end, or think-end) consistently reduces effectiveness. These comparisons confirm that pairing correctness-filtered teacher trajectories with the tail state yields the most informative target.
\subsection{Representation Analysis}

\begin{figure}[t]
\centering
\includegraphics[width=0.9\columnwidth]{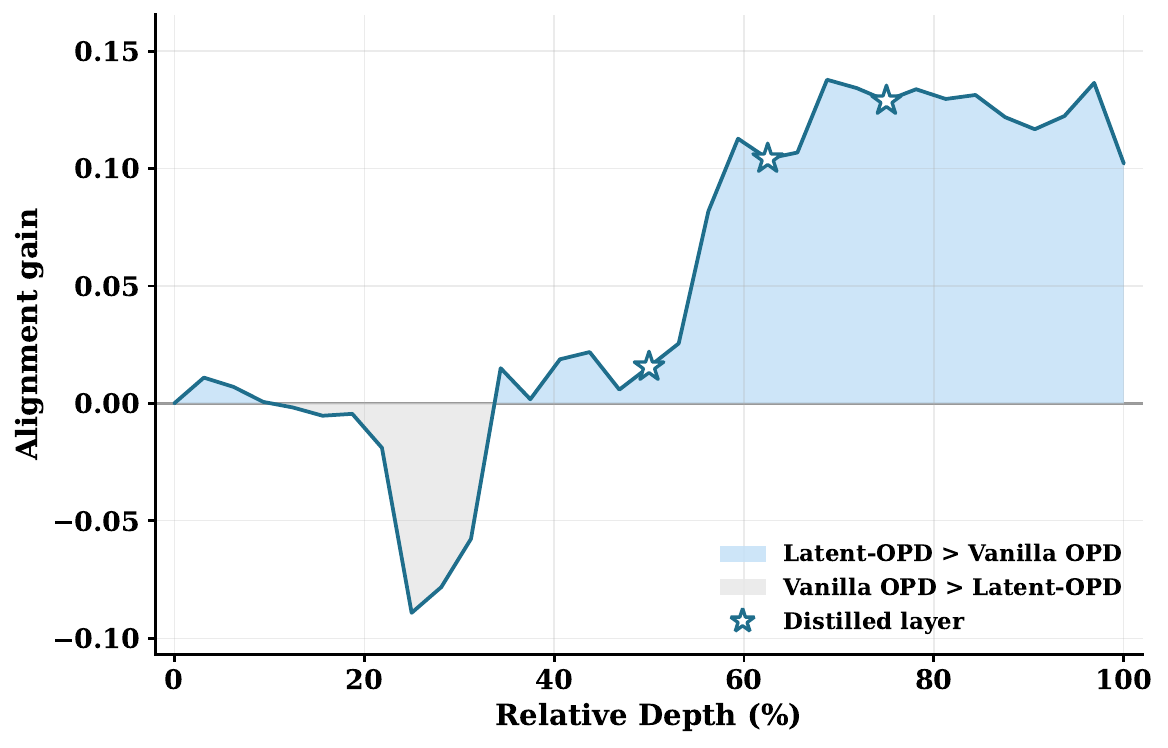}
\caption{Layer-wise teacher alignment shift ($\Delta\mathrm{CKA}$) from vanilla OPD to \textsc{Latent-OPD} on 16-frame Video-MME. Blue indicates increased similarity to the teacher, grey indicates decreased similarity, and stars mark directly supervised layers.}\label{fig:cka}
\vspace{-1em}
\end{figure}

\begin{figure*}[t]
\centering
\includegraphics[width=0.95\textwidth]{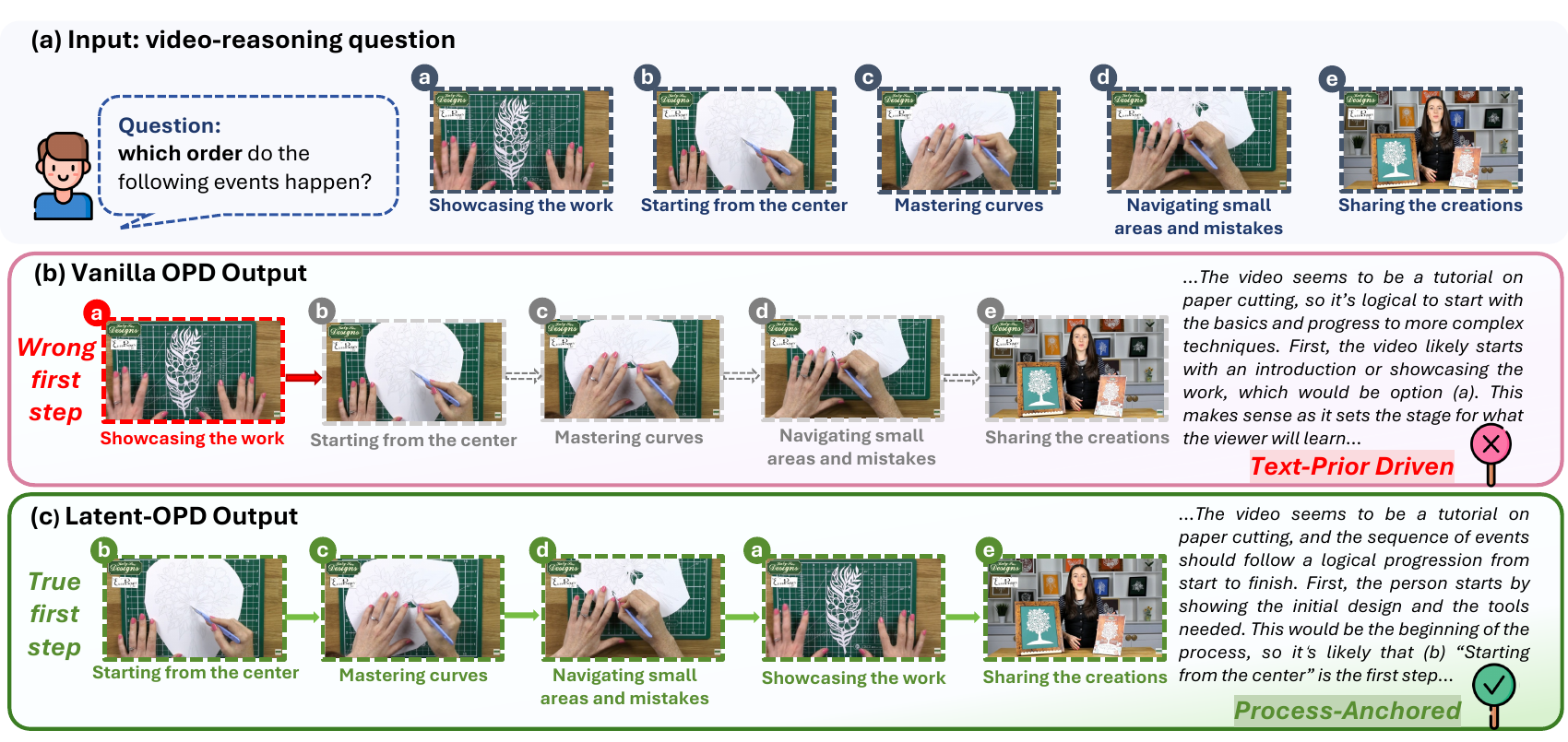}
\vspace{-0.5em}
\caption{Video-MME case study requiring the temporal ordering of five handicraft events (a). Vanilla OPD fails under all three frame budgets (b), whereas \textsc{Latent-OPD} consistently recovers the correct event order (c).}
\label{fig:case-study}
\vspace{-1em}
\end{figure*}
 To analyze internal representations, we compute projector-free linear Centered Kernel Alignment (CKA) on 16-frame Video-MME inference examples, which measures how similarly two layers organize the same set of examples, with higher values indicating more similar representation geometry. We compare vanilla OPD and \textsc{Latent-OPD} against the frozen teacher using one-to-one relative-depth matching, pairing each student layer with the teacher layer at the corresponding depth percentage. As shown in Figure~\ref{fig:cka}, the improvements are concentrated in the deep student layers, achieving an average CKA increase of $+0.108$ over the deep half and a peak gain of $+0.137$, while shallow layers remain nearly unchanged. This indicates that \textsc{Latent-OPD} mainly moves high-level pre-verbal reasoning states closer to the teacher, rather than broadly perturbing low-level early visual encoding. More details of the CKA analysis are provided in Appendix F.

\subsection{Case Study}

We further inspect a representative Video-MME example in Figure~\ref{fig:case-study}. The question asks the model to recover the order of five events in a handicraft video: starting from the center, mastering curves, navigating small areas and mistakes, showcasing the work, and sharing the creations. This is a procedural ordering problem, so the correct answer depends on preserving the event chain rather than recognizing a single salient frame. Vanilla OPD predicts incorrect options under all three frame budgets, and its mistakes are plausible language-prior shortcuts, e.g., placing ``showcasing the work'' too early. In contrast, \textsc{Latent-OPD} predicts the correct order $(b)(c)(d)(a)(e)$ consistently under all frames. While this case is not intended as standalone evidence, it illustrates the behavior suggested by the quantitative analysis: aligning to a correct teacher trajectory can help the student form a more faithful internal state for process-level video reasoning.

\section{Related Work}

\textbf{Visual and Video On-Policy Distillation.} On-policy distillation has been studied for language-model alignment and compression~\cite{minillm,opd}, and has recently been extended to multimodal settings. Vision-OPD~\cite{visionopd} and VA-OPD~\cite{liu2026visualadvantageonpolicydistillationvisionlanguage} introduce fine-grained visual supervision and vision-critical token reweighting; subsequent methods transfer text reasoning~\cite{vold}, steer visual grounding~\cite{decomposedopd,vicur}, or stabilize MLLM reasoning distillation~\cite{gndpo}. In the temporal domain, Video-OPD~\cite{videoopd} provides dense supervision for video grounding, while reasoning frameworks and rationale- or evidence-based video distillation transfer temporal reasoning only through generated outputs~\cite{videoofthought,aotd,qwen3tr,wu2026roboalign}.

\textbf{Latent and Representation-Level Reasoning.} Output logits are a compressed interface and may discard structural dynamics formed before verbalization~\cite{hao2025traininglargelanguagemodels,liu2026tangrampuzzle,deng2025survey,kuang2025express,li2024towards,liu2022we,zhang2026latentvisualcachevideo,kang2026lutlatentutilitytraining,xia2026perceptionreasoningdynamiclatent,adhikari2026multimodal}. Multimodal distillation studies show knowledge is encoded in intermediate hidden states~\cite{llavadi,ethink,li2022past,MatchLM2LiteAS}, while latent visual reasoning suggests internal representations support reasoning beyond explicit tokens~\cite{latentvisualreasoning}. OPRD~\cite{oprd} extends OPD into representation space for text LLMs.

 Directly applying OPRD to video is sub-optimal, as dense same-depth matching suffers from redundant frames and mismatched reasoning. \textsc{Latent-OPD} instead aligns compact trajectory-tail states after cross-frame integration and maps middle-to-late student layers to deeper teacher layers, adapting representation-level OPD for spatiotemporal reasoning.

\section{Conclusion}

In this work, we demonstrate that effective video reasoning depends not only on output-token preference matching but also on the latent spatiotemporal states that organize sparse visual evidence before it is verbalized. We propose \textsc{Latent-OPD}, augmenting on-policy distillation with compact trajectory-tail hidden-state alignment and progressive teacher-lookahead mapping. Across six benchmarks, \textsc{Latent-OPD} outperforms vanilla OPD, especially under limited frame budgets. These results support deep thought alignment: distilling a video reasoner must go beyond final answer imitation to transfer teacher integration of events, relations, and temporal structure. By turning hidden reasoning trajectories into practical supervision, we hope \textsc{Latent-OPD} offers effective guidance for developing more frame-efficient and reliable LLMs.

\nocite{videollama2,llamavid,visd,vited,rethinkvideocot}

\bibliography{aaai2027}

\appendix
\section*{Appendix}

This appendix documents implementation details omitted from the main paper for space and reproducibility. Section~\ref{app:hyperparams} presents the hyperparameters and training setup of \textsc{Latent-OPD}, including the training and evaluation data, public baselines, model initialization, the two-stage post-training recipe, the output and latent objectives, teacher trajectory generation, the optimization schedule, and video preprocessing. Section~\ref{app:4b-results} reports additional 4B-scale experimental results across all three frame budgets, and Section~\ref{app:teacher-results} presents the Qwen3.5-27B teacher model results under the same evaluation protocol. Section~\ref{app:finegrained} provides fine-grained Video-MME analyses, including breakdowns by video duration and by domain. Section~\ref{app:design-details} details the ablation setting and the design choices tested in the main paper. Section~\ref{app:cka-details} reports the CKA protocol behind the one-to-one layer-alignment curve.

\section{Hyperparameters and Training Setup}
\label{app:hyperparams}

\subsubsection{Training and evaluation data}
All post-training data follows Video-R1~\cite{videor1}. Stage~I (SFT) uses the Video-R1-CoT-165k cold-start corpus, which annotates a mixed image--video pool with chain-of-thought rationales and removes low-quality or inconsistent generations. Stage~II (RL/OPD) samples from the corresponding Video-R1-260k post-training set. The video portion is intended to strengthen temporal understanding, including event order, frame-to-frame dependency, motion, and causal dynamics, while the image portion supplies complementary static reasoning supervision such as chart and OCR understanding, mathematical and spatial reasoning, and knowledge-intensive visual QA. Since most examples have verifiable multiple-choice or numerical answers, they support rule-based correctness checks for on-policy optimization and for retaining teacher trajectories in our latent branch. For evaluation, we follow the public Video-R1 benchmark protocol on VSI-Bench~\cite{vsibench}, Video-MMMU~\cite{videommmu}, MMVU~\cite{mmvu}, MVBench~\cite{mvbench}, TempCompass~\cite{tempcompass}, and Video-MME~\cite{videomme}. We report accuracy under the same $16$/$32$/$64$-frame input budgets and do not mix evaluation samples into training.

\subsubsection{Baselines}
The public baselines include LongVA-7B~\cite{longva}, VILA-1.5-8B and VILA-1.5-40B~\cite{vila}, Video-UTR-7B~\cite{videoutr}, LLaVA-OneVision-7B~\cite{llavaonevision}, Kangaroo-8B~\cite{kangaroo}, and Video-R1-7B~\cite{videor1}. We additionally implement Qwen3.5-9B$_{\text{CoT}}$, Qwen3.5-9B$_{\text{SFT+GRPO}}$, Qwen3.5-9B$_{\text{Vanilla OPD}}$, and Qwen3.5-9B$_{\text{Latent-OPD}}$ under the same frame-budget protocol.

\subsubsection{Models}
The student is initialized from the Qwen3.5-9B-Base video CoT checkpoint (text hidden size $4096$, $32$ text layers, full-attention interval $4$). The teacher is a Qwen3.5-27B video CoT SFT checkpoint obtained by fine-tuning Qwen3.5-27B-Base for $2{,}000$ steps (text hidden size $5120$, $64$ text layers, full-attention interval $4$). All layer indices in the teacher-lookahead pairs are snapped to multiples of $4$ to align with full-attention blocks.

\subsubsection{Two-stage post-training}
Stage~I is SFT on the Video-R1 CoT corpus. Stage~II runs on-policy distillation, on top of which \textsc{Latent-OPD} adds a teacher-trajectory hidden-state alignment auxiliary loss. The OPD objective and the latent objective share the same input batch but use different trajectories: OPD follows student rollouts, while latent alignment uses retained teacher trajectories. Their gradients are accumulated separately, and the latent loss is applied as an auxiliary backpropagation term rather than a policy-gradient reward.

\subsubsection{Output objective}
JSD-style on-policy distillation with mixing weight $\alpha=0.5$, applied on the student-selected top-$100$ token support plus an aggregated tail bucket. Reference KL weight $\beta=0.04$. The format regularization weight is $\lambda_{\text{fmt}}=0.05$. Rollout uses $4$ sampled completions per prompt with a generation batch size of $8$.

\subsubsection{Latent objective}
Correct-only filtering with correctness threshold $1.0$ and default-keep when no label is available. Three teacher-lookahead layer pairs: $(s_{16},t_{48})$, $(s_{20},t_{56})$, $(s_{24},t_{64})$, corresponding to relative depths $(50\%,75\%)$, $(62.5\%,87.5\%)$, $(75\%,100\%)$. The latent weight is $\lambda_g=0.01$, with SFT- and prompt-end latent branches disabled. The auxiliary loss is computed in cosine distance after a full-rank cross-architecture projector (per pair $\approx 20.97$M parameters, three pairs $\approx 62.9$M parameters), normalized along the hidden dimension, and capped at $15\%$ of the generative loss under a scaling cap. A linear warm-up of $5\%$ of total steps is applied to the latent weight.

\subsubsection{Teacher trajectory generation}
Done on the fly within the training loop with up to $512$ new tokens, sampling temperature $0.7$ and top-$p$ $0.9$, with an answer-close early-stopping rule. Each batch reuses the same teacher trajectory for both student and teacher forward passes, and all teacher parameters are frozen under \texttt{no\_grad}.

\subsubsection{Optimization and schedule}
Stage~I uses AdamW with weight decay and a cosine learning-rate schedule. Stage~II uses the same optimizer with $\beta_1=0.9$, $\beta_2=0.95$, no decay on bias and LayerNorm, and gradient clipping at $1.0$. The student is trained for $300$ steps in total for the main experiment, with checkpoints saved every $100$ steps.

\subsubsection{Video preprocessing} Training clips are sampled at $1$ FPS and capped at $16$ frames. Visual inputs are tokenized with a patch size of $28\times 28$. Each example's prompt and video are pre-tokenized once and re-used across the four rollouts.

\begin{table*}[t]
\centering
\footnotesize
\setlength{\tabcolsep}{5pt}
\renewcommand{\arraystretch}{1.12}

\resizebox{\textwidth}{!}{
\begin{tabular}{l|c|cccccc|c}
\toprule[1.0pt]

\textbf{Model} & \textbf{Frames} & \textbf{VSI-Bench} & \textbf{Video-MMMU} & \textbf{MMVU} & \textbf{MVBench} & \textbf{TempCompass} & \textbf{Video-MME} & \textbf{Avg.} \\
\midrule
 \textnormal{Vanilla OPD} & 16 & 44.5 & 59.8 & 68.0 & \textbf{64.7} & 73.4 & 56.0 & 61.1 \\
 \textnormal{Vanilla OPD} & 32 & 46.8 & 60.9 & 68.2 & \textbf{65.5} & \textbf{73.7} & 60.3 & 62.6 \\
 \textnormal{Vanilla OPD} & 64 & \textbf{50.2} & 61.3 & 69.0 & \textbf{66.8} & 74.0 & 63.3 & 64.1 \\
 \textsc{Latent-OPD} & 16 & \textbf{45.1} & \textbf{64.6} & \textbf{69.4} & 64.5 & \textbf{73.5} & \textbf{57.7} & \textbf{62.5} \\
 \textsc{Latent-OPD} & 32 & \textbf{49.0} & \textbf{66.3} & \textbf{69.6} & 65.3 & 73.2 & \textbf{61.3} & \textbf{64.1} \\
 \textsc{Latent-OPD} & 64 & 49.6 & \textbf{65.1} & \textbf{71.7} & 66.2 & \textbf{74.1} & \textbf{64.9} & \textbf{65.3} \\
\bottomrule[1.0pt]
\end{tabular}
}

\caption{Qwen3.5-4B accuracy (\%) across six benchmarks and three frame budgets. Bold marks the better result between Vanilla OPD and \textsc{Latent-OPD} for each benchmark--budget pair.}
\label{tab:app-4b-full}
\end{table*}
\begin{table*}[tp]
\centering
\footnotesize
\setlength{\tabcolsep}{5pt}
\renewcommand{\arraystretch}{1.12}

\resizebox{\textwidth}{!}{
\begin{tabular}{l|c|cccccc|c}
\toprule[1.0pt]

\textbf{Model} & \textbf{Frames} & \textbf{VSI-Bench} & \textbf{Video-MMMU} & \textbf{MMVU} & \textbf{MVBench} & \textbf{TempCompass} & \textbf{Video-MME} & \textbf{Avg.} \\
\midrule
 Teacher 27B & 16 & 49.0 & 71.7 & 75.5 & 65.4 & \textbf{74.1} & 63.9 & 66.6 \\
Teacher 27B & 32 & 52.2 & \textbf{74.0} & 75.0 & 65.9 & 73.5 & 67.1 & 68.0 \\
Teacher 27B & 64 & \textbf{54.1} & 73.0 & \textbf{76.6} & \textbf{66.6} & 72.8 & \textbf{69.6} & \textbf{68.8} \\
\bottomrule[1.0pt]
\end{tabular}
}

\caption{Qwen3.5-27B teacher accuracy (\%) on the six benchmarks under the same 16-, 32-, and 64-frame protocol. The teacher is obtained after $2{,}000$ SFT steps; bold marks the best frame budget for each benchmark.}
\label{tab:teacher-results}
\end{table*}

\section{Additional 4B-Scale Results}
\label{app:4b-results}

Beyond the 9B setting, Figure~\ref{fig:4b-f16} presents the 16-frame comparison for a Qwen3.5-4B student, and Table~\ref{tab:app-4b-full} extends it to all three frame budgets. The 4B student follows the same evaluation protocol as the 9B experiments. \textsc{Latent-OPD} improves the average score consistently, from $61.1\%$ to $62.5\%$ at 16 frames, from $62.6\%$ to $64.1\%$ at 32 frames, and from $64.1\%$ to $65.3\%$ at 64 frames. The improvements on Video-MMMU, MMVU, and Video-MME indicate that the trajectory-level latent signal remains effective even when the student capacity is further reduced.

\section{Teacher Model Results}
\label{app:teacher-results}

For context, Table~\ref{tab:teacher-results} reports the Qwen3.5-27B teacher checkpoint under the same six-benchmark, three-frame evaluation protocol. This teacher is not an off-the-shelf base model; it is obtained by fine-tuning Qwen3.5-27B-Base for $2{,}000$ steps on the video CoT SFT data.

Compared with the 9B \textsc{Latent-OPD} student in Table~\ref{tab:main-results}, the teacher remains stronger on most benchmarks, yet the gap is moderate. Averaged over the six benchmarks, the teacher reaches $66.6\%$, $68.0\%$, and $68.8\%$ at 16, 32, and 64 frames, while the \textsc{Latent-OPD} student reaches approximately $64.4\%$, $65.6\%$, and $66.5\%$ under the same frame budgets. The remaining average gaps are therefore about $2.2$, $2.4$, and $2.3$ points. The largest residual gap appears on Video-MMMU, where the 27B teacher benefits from stronger knowledge capacity; by contrast, the student is much closer on VSI-Bench, MVBench, TempCompass, and Video-MME. At 64 frames, the student even slightly exceeds the teacher on VSI-Bench ($54.9\%$ vs. $54.1\%$), suggesting that \textsc{Latent-OPD} transfers much of the teacher's video-reasoning behavior while retaining efficient 9B-scale inference.

\section{Video-MME Fine-Grained Results}
\label{app:finegrained}

To complement the aggregate analysis in Section~\ref{sec:main-results}, we report the full Video-MME breakdowns on the 2700-question validation split. The results use the same checkpoints and decoding protocol as the experiments, and compare vanilla OPD with \textsc{Latent-OPD} under 16, 32, and 64 input frames.

\begin{figure*}[t]
\centering
\includegraphics[width=0.92\textwidth]{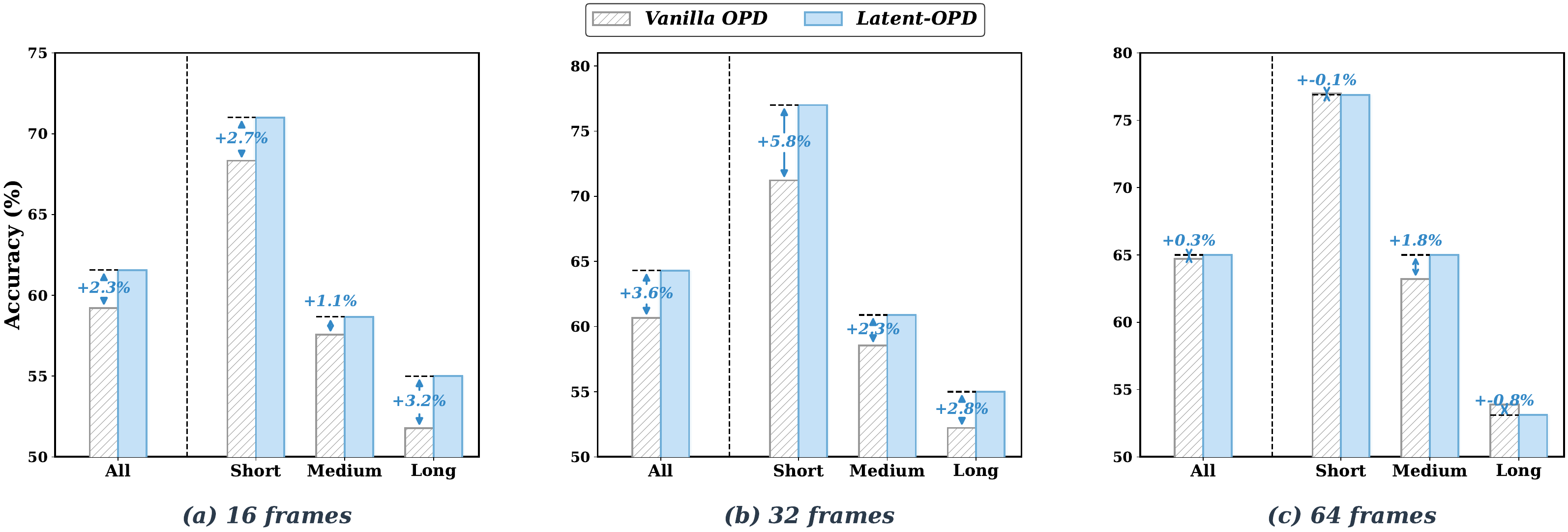}
\caption{Video-MME accuracy (\%) by video duration under 16-, 32-, and 64-frame budgets. \textsc{Latent-OPD} shows its largest long-video gain in the 16-frame setting.}
\label{fig:app-videomme-duration}
\end{figure*}

\begin{figure*}[t]
\centering
\includegraphics[width=0.92\textwidth]{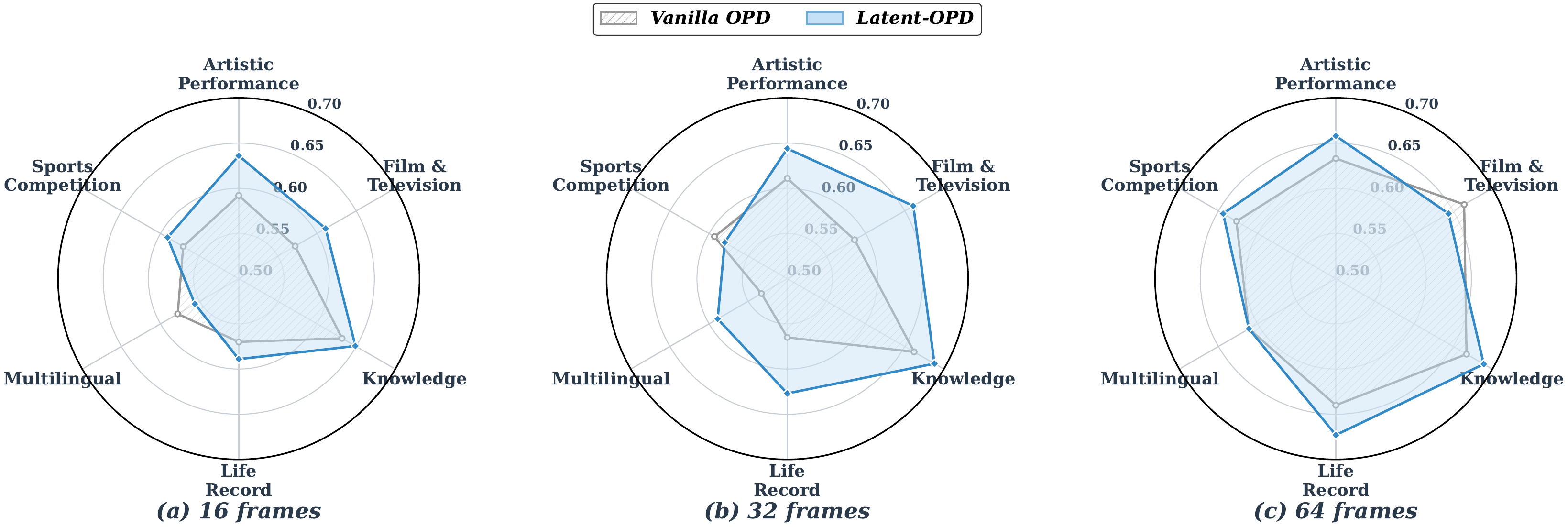}
\caption{Video-MME accuracy (\%) by domain under 16-, 32-, and 64-frame budgets. The largest low- and mid-frame gains occur in Film \& Television, Artistic Performance, and Life Record.}
\label{fig:app-videomme-domain}
\end{figure*}

\begin{table}[t]
\centering
\footnotesize
\setlength{\tabcolsep}{2.2pt}
\renewcommand{\arraystretch}{1.14}

\begin{tabularx}{\columnwidth}{@{}l@{\hspace{0.7em}}l@{\hspace{0.7em}}>{\centering\arraybackslash}X>{\centering\arraybackslash}X>{\centering\arraybackslash}X@{}}
\toprule[1.0pt]

\textbf{Duration} & \textbf{Model} & \textbf{16} & \textbf{32} & \textbf{64} \\
\midrule
Short & Vanilla OPD & 68.33 & 71.22 & \textbf{77.00} \\
 & \textsc{Latent-OPD} & \textbf{71.00} & \textbf{77.00} & 76.89 \\
Medium & Vanilla OPD & 57.56 & 58.56 & 63.22 \\
 & \textsc{Latent-OPD} & \textbf{58.67} & \textbf{60.89} & \textbf{65.00} \\
Long & Vanilla OPD & 51.78 & 52.22 & \textbf{53.89} \\
 & \textsc{Latent-OPD} & \textbf{55.00} & \textbf{55.00} & 53.11 \\
\bottomrule[1.0pt]
\end{tabularx}

\caption{Video-MME accuracy (\%) by video duration and frame budget. Bold marks the higher result between vanilla OPD and \textsc{Latent-OPD} in each setting.}
\label{tab:app-duration}
\end{table}

\begin{table}[t]
\centering
\scriptsize
\setlength{\tabcolsep}{1.5pt}
\renewcommand{\arraystretch}{1.12}

\begin{tabularx}{\columnwidth}{@{}>{\raggedright\arraybackslash}p{0.28\columnwidth}@{\hspace{0.35em}}>{\raggedright\arraybackslash}p{0.23\columnwidth}@{\hspace{0.35em}}>{\centering\arraybackslash}X>{\centering\arraybackslash}X>{\centering\arraybackslash}X@{}}
\toprule[1.0pt]

\textbf{Domain} & \textbf{Model} & \textbf{16} & \textbf{32} & \textbf{64} \\
\midrule
Life Record & Vanilla OPD & 56.98 & 56.51 & 63.97 \\
 & \textsc{Latent-OPD} & \textbf{58.89} & \textbf{62.70} & \textbf{65.87} \\
Film \& Television & Vanilla OPD & 57.22 & 58.61 & \textbf{66.39} \\
 & \textsc{Latent-OPD} & \textbf{61.11} & \textbf{66.11} & 64.44 \\
Artistic Performance & Vanilla OPD & 59.17 & 61.11 & 63.33 \\
 & \textsc{Latent-OPD} & \textbf{63.61} & \textbf{64.44} & \textbf{64.72} \\
Knowledge & Vanilla OPD & 63.21 & 66.17 & 66.67 \\
 & \textsc{Latent-OPD} & \textbf{64.94} & \textbf{68.77} & \textbf{67.65} \\
Multilingual & Vanilla OPD & \textbf{57.78} & 53.33 & \textbf{61.11} \\
 & \textsc{Latent-OPD} & 55.56 & \textbf{58.89} & 58.89 \\
Sports Competition & Vanilla OPD & 57.11 & \textbf{59.33} & \textbf{62.67} \\
 & \textsc{Latent-OPD} & \textbf{59.11} & 58.00 & 60.89 \\
\bottomrule[1.0pt]
\end{tabularx}

\caption{Video-MME accuracy (\%) by domain and frame budget. Bold marks the higher result between vanilla OPD and \textsc{Latent-OPD} in each setting.}
\label{tab:app-domain}
\end{table}

Taken together, Figures~\ref{fig:app-videomme-duration}--\ref{fig:app-videomme-domain} and Tables~\ref{tab:app-duration}--\ref{tab:app-domain} make the same pattern visible from complementary angles. \textsc{Latent-OPD} is most helpful when the student must compress sparse or temporally distributed visual evidence into a global reasoning state: it improves long videos under 16 frames, strengthens process-oriented domains such as Film \& Television and Artistic Performance. Meanwhile, the smaller or mixed gains at 64 frames clarify the boundary condition of the method. When visual evidence is already dense, vanilla OPD can recover part of the missing information simply by observing more frames, so the marginal value of latent trajectory anchors naturally becomes smaller.

\section{Ablation Setting Details}
\label{app:design-details}

Table~\ref{tab:ablation} isolates one design choice at a time under a shared training and evaluation recipe. Unless stated otherwise, all variants start from the same 9B SFT initialization, use the Video-R1 post-training data, train for $300$ steps with $4$ sampled rollouts per prompt, use $16$ training frames, and evaluate on Video-MME with $16$/$32$/$64$ input frames. The output objective is the same JSD-style OPD loss used in the main method: it is evaluated on the student-selected top-$100$ support plus an aggregated tail bucket, together with reference-policy and format regularization. vanilla OPD disables only the latent branch, whereas \textsc{Latent-OPD} adds correctness-filtered teacher trajectories, the last non-padding response token anchor, full-rank projectors, and the default teacher-lookahead pairs $(s_{50\%}\!\to\!t_{75\%})$, $(s_{62.5\%}\!\to\!t_{87.5\%})$, and $(s_{75\%}\!\to\!t_{100\%})$.

\textbf{Controlled layer mapping variants.} The layer-mapping group fixes the student depths at $50\%$, $62.5\%$, and $75\%$ and varies only the teacher side. This keeps the number of aligned states, projector sizes, latent weight, correctness filter, and loss cap unchanged, so the variants directly test how the teacher depth paired with each student state affects the auxiliary signal. Same-depth uses $(s_{50\%}\!\to\!t_{50\%})$, $(s_{62.5\%}\!\to\!t_{62.5\%})$, and $(s_{75\%}\!\to\!t_{75\%})$; fixed-offset uses $(s_{50\%}\!\to\!t_{62.5\%})$, $(s_{62.5\%}\!\to\!t_{75\%})$, and $(s_{75\%}\!\to\!t_{87.5\%})$; and reverse-lookahead uses $(s_{50\%}\!\to\!t_{25\%})$, $(s_{62.5\%}\!\to\!t_{50\%})$, and $(s_{75\%}\!\to\!t_{62.5\%})$. We also include pair-count controls: the single-tail variant keeps only $(s_{75\%}\!\to\!t_{100\%})$, the final-student variant adds $(s_{100\%}\!\to\!t_{100\%})$, and the early-pair variant adds $(s_{25\%}\!\to\!t_{50\%})$ to the default mapping. Together, these controls separate the effect of teacher lookahead from simply adding more projectors or moving supervision to the final student layer.

\textbf{Representation-level and token-level controls.} The OPRD-style baseline adapts representation-level OPD to this setting by applying latent supervision on student on-policy rollouts, matching all response-token hidden states densely, using same-depth pairs, and optimizing normalized MSE. The dense token-level hidden KD variant keeps teacher trajectories but aligns response-token hidden states rather than the single tail anchor, testing whether denser hidden matching is actually helpful for video. The low-rank projector variant replaces each full-rank $4096\!\to\!5120$ projector with a rank-$16$ bottleneck while leaving the trajectory source and layer pairs unchanged. The teacher-trajectory SFT baseline uses the same correctness-filtered teacher trajectories as \textsc{Latent-OPD} but replaces hidden-state alignment with next-token CE on the teacher response, masking all tokens when the teacher's final answer is incorrect. Finally, the Reverse-KL output-distillation variant changes the output objective while keeping the latent branch comparable. These controls distinguish the proposed sparse latent alignment from dense feature matching, projector-capacity reduction, simple exposure to correct teacher rationales, and a nearby output-level divergence.

\textbf{Trajectory and anchor variants.} The source controls test whether the latent target is both reliable and position-matched. The unfiltered variant feeds the same teacher-generated trajectory to both models but keeps incorrect teacher answers. The independent-trajectory variant lets teacher and student generate separate completions before aligning their tail states, so the two states can summarize different reasoning paths. The shared-student-trajectory variant restores positional correspondence by feeding the same student completion to both models, but the teacher state is then conditioned on a path the teacher did not choose. \textsc{Latent-OPD} instead feeds the identical correctness-filtered teacher trajectory to both models, combining positional correspondence with the teacher's preferred correct path. Anchor variants then replace the default last non-padding response token with prompt-end, answer-end, or think-end states. The default anchor is designed to learn the complete response-state summary rather than an intermediate prompt or thinking-state representation: after the teacher has read the visual evidence, constructed the reasoning chain, and reached the end of the valid response, this hidden state compactly summarizes the question semantics, spatial-temporal evidence, action and event ordering, and the reasoning context that leads to the final decision. In contrast, the prompt-end state is collected before the response is formed, the answer-end state may reflect only answer selection, and the think-end state may miss the final consolidation from reasoning to answer. Their lower averages therefore support the default design: correctness-filtered teacher trajectories are most useful when distilled through the final valid response state, where the model has already integrated the video, question, and generated reasoning context into a global summary representation.

\section{Additional CKA Analysis}
\label{app:cka-details}

To support the representation analysis, we report the CKA protocol used to produce the one-to-one layer-alignment curve. Let $H^{\mathrm{v}}_{s}$ and $H^{\mathrm{o}}_{s}$ denote the trajectory-tail hidden states of vanilla OPD and our method at student layer $s$, and let $T_t$ denote the teacher states at layer $t$ on the same $N$ examples. After centering examples, projector-free linear CKA is computed from sample Gram matrices:
\begin{equation}
\operatorname{CKA}(H_s,T_t)=
\frac{\langle \tilde H_s\tilde H_s^{\top},\tilde T_t\tilde T_t^{\top}\rangle_F}
{\|\tilde H_s\tilde H_s^{\top}\|_F\,\|\tilde T_t\tilde T_t^{\top}\|_F+\epsilon}.
\end{equation}
Because this score depends only on example-example similarities, it can compare the 4096-dimensional student states with the 5120-dimensional teacher states without requiring a learned projector. We avoid max-pooling over teacher layers and match each student layer $s$ to the teacher layer at the same relative depth:
\begin{equation}
t(s)=\operatorname{round}(64s/32).
\end{equation}
The plotted gain is $\Delta\operatorname{CKA}(s)=\operatorname{CKA}(H^{\mathrm{o}}_{s},T_{t(s)})-\operatorname{CKA}(H^{\mathrm{v}}_{s},T_{t(s)})$, with stars marking the directly supervised student layers $s\in\{16,20,24\}$.

This depth-aligned design asks a direct question: at the same relative depth, is the student representation closer to the teacher after latent alignment? The answer is concentrated in the deeper half of the student. As shown in Figure~\ref{fig:cka}, \textsc{Latent-OPD} produces an average CKA gain of about $+0.108$ over student layers at or beyond $50\%$ depth, with a peak gain of about $+0.137$, while shallow layers remain nearly unchanged. This pattern matches the training design: the supervised layers are in the middle-to-late region, and the latent objective is applied only to compact trajectory-tail states. Thus, the auxiliary latent objective does not broadly perturb local visual encodings; it moves high-level pre-verbal reasoning states closer to the teacher manifold.

\end{document}